\pdfoutput=1

\documentclass[11pt]{article}

\usepackage{acl}

\usepackage{times}
\usepackage{latexsym}
\usepackage{bm}
\usepackage{amsmath}
\usepackage{amssymb}
\usepackage{caption}
\usepackage{subcaption}
\usepackage{booktabs}
\usepackage{url}
\usepackage[export]{adjustbox}
\usepackage[ruled]{algorithm2e}

\SetCommentSty{mycommfont}
\usepackage{algorithmic}

\usepackage[T1]{fontenc}

\usepackage[utf8]{inputenc}

\usepackage{microtype}

\usepackage{inconsolata}

\newcommand{\method}{\texttt{ATP}}

\newcommand{\red}[1]{\textcolor{red}{#1}}

\newcommand{\blue}[1]{\textcolor{blue}{#1}}

\newtheorem{theorem}{Theorem}[section]

\newtheorem{remark}[theorem]{Remark}

\newcommand{\ui}[1]{\bm{u}_{#1}}
\newcommand{\vi}[1]{\bm{v}_{#1}}

\usepackage{tikz}
\usepackage{pgfplots}
\usepackage{environ}
\pgfplotsset{compat = newest}
\usetikzlibrary{arrows,positioning} 
\usetikzlibrary{decorations.text}
\usetikzlibrary{decorations.pathreplacing}
\usetikzlibrary{arrows.meta}
\tikzset{
    >=stealth',
    port/.style = {circle, draw, align=center, minimum height=1mm},
    op/.style={
           rectangle,
           rounded corners,
           draw=black, thick,
           text width=3.5em,
           minimum height=1em,
           text centered},
    data/.style={
           rectangle,
           draw=black, thick,
           minimum height=1em,
           text centered},
    area/.style={
           rectangle,
           draw=black, thick,
           minimum height=1em},
    connect/.style={
           ->,
           thick,
           shorten <=2pt,
           shorten >=2pt,}
}

\title{\method{}: Enabling Fast LLM Serving via Attention on Top Principal Keys  }

\author{Yue Niu \\
  {\small University of Southern California} \\
  {\small Los Angeles, CA, US} \\
  \texttt{yueniu@usc.edu} \\\And
  Saurav Prakash \\
  {\small University of Illinois Urbana-Champaign} \\
  {\small Urbana, IL, US} \\
  \texttt{sauravp2@illinois.edu} \\\And
  Salman Avestimehr \\
  {\small University of Southern California} \\
  {\small Los Angeles, CA, US} \\
  \texttt{avestime@usc.edu}
}

\begin{document}
\maketitle
\begin{abstract}
We propose a new attention mechanism with linear complexity, \method{}, that fixates \textbf{A}ttention on \textbf{T}op \textbf{P}rincipal keys, rather than on each individual token. Particularly, 
\method{} is driven by an important observation that input sequences are typically low-rank, i.e., input sequences can be represented by a few principal bases.
Therefore, instead of directly iterating over all the input tokens, \method{} transforms inputs into an orthogonal space and computes attention only on the top principal bases (keys).
Owing to the observed low-rank structure in input sequences, \method{} is able to capture semantic relationships in input sequences with a few principal keys.
Furthermore, the attention complexity is reduced from \emph{quadratic} to \emph{linear} without incurring a noticeable performance drop. 
\method{} further reduces complexity for other linear layers with low-rank inputs, leading to more speedup compared to prior works that solely target the attention module.
Our evaluations on various models (e.g., BERT and Llama) demonstrate that \method{} achieves comparable accuracy with much lower computation and memory complexity than the standard attention mechanism. In particular, \method{} barely loses accuracy with only $1/2$ principal keys, and only incurs around $2\%$ accuracy drops with $1/4$ principal keys. 
\end{abstract}
\section{Introduction}\label{sec:intro}
Transformers with self-attention have become a mainstream model architecture in many machine-learning tasks on natural language processing \cite{TransformersNLP}, and computer vision \cite{TransformersVision, VIT_ICLR_2020}. 
In particular, owing to the attention mechanism, transformers have been demonstrated to be more effective in learning semantic relationships from input sequences. 
This drives transformers to become the backbone of current large language models (LLMs) like ChatGPT \cite{ChatGPT} and Copilot \cite{Copilot}. 

Despite their remarkable utility in real-world applications, transformers with standard self-attention, however, incur \emph{quadratic} complexity in terms of sequence length \citep{Transformer_NIPS_2017}. 
To be specific, considering an input sequence with length $L$ (i.e., $L$ tokens), each attention layer needs $\mathcal{O}(L^2)$ computation and memory complexity on attention operations.
Such a quadratic degree of complexity renders transformers difficult to scale with long input sequences.  
As a result, most LLM services at scale backed by transformers incur significant computation and memory footprints, which can only be afforded by large companies with sufficient computing power \cite{LLMCosts_2023_arXiv}. 
To meet memory and computation resource constraints during deployment, some transformer models \cite{BERT_2018_arXiv, ALBERT_2020_ICLR, GPT_2018_OpenAI, Llama_2023_arXiv} usually come with a hard constraint on sequence length.
However, in many real-world tasks such as question-answering \cite{Superglue_NIPS_2019}, text summarization \cite{TextSummary_ELS_2021}, enabling long sequence length is crucial for capturing semantic relationships in a broader context, and improving models' performance. 

Therefore, at the core of transformers and LLM services, lightweight self-attention mechanisms play a key role in improving model performance with longer sequences, as well as computation and memory efficiency in deployment.

Current works on reducing the complexities of transformers can be categorized in two ways. The first line of research usually exploits redundancy in query/key/value matrices or attention maps, while the second approximates the \texttt{Softmax}-based attention with linear complexity. \\
Along the first line of works, \citet{Clustering_NIPS_2020} reduces attention complexity via clustering queries and only computes attention output for each cluster.
Its performance hinges on the performance of clustering as well as the dimension in queries. 
On the other hand, Linformer \cite{LinFormer_arXiv_2020} chooses to reduce the number of keys/values via a low-dimensional projection. 
A pre-defined or learnable projection layer is inserted into each attention layer.
However, such a projection layer lacks a rigorous guarantee to preserve information in inputs. \\
Compared to simply approximating queries, keys, or values, another line of work approximates the \texttt{Softmax}-based attention using randomized feature mapping\cite{RandomFeature_NIPS_2007}. 
In these works, standard attention is regarded as a kernel method, and can be approximated with low-dimensional kernels. 
For instance, Performers \citep{Performers_ICLR_2021} shows that \texttt{Softmax} attention can be converted to a \texttt{Gaussian} kernel function.
Therefore, self-attention can be potentially calculated at a lower dimension with linear complexity, as done in a \texttt{Gaussian} kernel function.
Following such an idea, several works explore different kernels to approximate \texttt{Softmax} attention \cite{LinearFormer_ICML_2020}.
However, the approximation methods with randomized feature mapping need to trade off approximation accuracy and the number of random features.
A low approximation error needs more random features but increases approximation complexity \cite{Performers_ICLR_2021}.
Furthermore, while the aforementioned works reduce the complexity of the attention mechanism to linear, they still have not been seen in large-scale language models such as Llama \cite{Llama_2023_arXiv}. One reason is that these methods cannot effectively preserve information when performing attention in low dimensions.  

In this paper, we propose a new attention mechanism with linear complexity, that maintains model performance with significantly reduced computation and memory costs.  
The new attention mechanism, called \method{}, is the first work that adapts self-attention with a low-rank structure in input embeddings. 
In particular, \method{} first analyzes an input sequence's structure and obtains orthogonal bases. 
We observe that input sequences usually exhibit a high correlation among tokens, with a few orthogonal bases being more important than the rest.
Then \method{} computes attention only on the top principal bases ( we call them \emph{principal keys}), rather than iterating over all keys as in the standard attention layer.
Owing to the low-rank structure in input sequences, the new self-attention mechanism with few principal keys/values is sufficient to capture semantic relationships among tokens. 
As a result, compute and memory complexity is reduced from quadratic to linear in terms of sequence length. \\
Furthermore, by exploiting low-rank structure in inputs, not only is the complexity of attention reduced, but also the complexity of other linear layers.
Hence, \method{} achieves further computation reductions compared to prior works focusing solely on the \texttt{Softmax} operation.

Our evaluations on various models (e.g., BERT and Llama) demonstrate 
\method{} still maintains comparable accuracy with small fractional principal keys/values.
In particular, with only $1/4$ principal keys, \method{} achieves accuracy almost as the original model. With only $1/4$ principal keys, \method{} only incurs around $2\%$ accuracy drop on BERT-base and Llama2 models. 

\section{Preliminaries and Related Works}\label{sec:prem}
\subsection{Standard Self-Attention}\label{sec:prem:att}
Standard self-attention consists of three matrices: queries $Q$, keys $K$, and values $V \in \mathbb{R}^{L\times d'}$, where $L$ is sequence length and $d'$ is the hidden dimension.  
For each query vector $\bm{q} \in Q$, the self-attention applies dot-product with all keys, followed by a \texttt{Softmax} op to compute a score on each key. 
Each score denotes a weight on the corresponding values. Then the attention output $A(\bm{q})$ is obtained as a weighted average of all values:
 
\begin{equation}\label{eq:att}
    A(\bm{q}) = \texttt{Softmax}(\bm{q} \cdot K^T / \sqrt{d}) \cdot V.
\end{equation}

The query/key/value matrices are obtained by projecting input $X \in \mathbb{R}^{L\times d}$ with parameter $W^Q, W^K, W^V \in \mathbb{R}^{d\times d'}$ as

\begin{equation}
    Q, K, V = X \cdot \{ W^Q, W^K, W^V \}.
\end{equation}

In essence, the self-attention mechanism finds the relations between a query and all keys, which are measured by probability after \texttt{Softmax}. Then, it averages corresponding values with the notion that a key closer to the query should be assigned larger weights (i.e., probability after \texttt{Softmax}).  

\subsection{Related Works on Efficient Self-Attention}\label{subsec:related}
Given an input sequence with length $L$,  the standard self-attention needs to perform $L^2$ dot-products for all token vectors to get the whole attention map.
As a result, it incurs complexity of $\mathcal{O}(L^2)$ on computations and memory, which makes it difficult to scale with long inputs in many tasks. 
As a result, current LLM services usually require a significant amount of memory and computing power, in order to support long sequences. 
In some cases, facing actual resource constraints, some LLMs may need to limit the sequence length. 

Current literature typically mitigates the limitation via exploiting sparsity or redundancy in attention matrices \cite{SparseAtt_ACL_2021,SparseAtt_ICLR_2021,Clustering_NIPS_2020,LinearFormer_ICML_2020,LinFormer_arXiv_2020}, or approximating the self-attention operation. 
For sparsity and redundancy in attention, exploiting low-rank structures in query/key/value and attention maps shows great potential. For instance, by exploring redundancy in input query vectors, \citet{Clustering_NIPS_2020} propose to first cluster query vectors, and use \emph{cluster centroid vectors} to represent all query vectors of the same cluster. Hence, for all queries in the same cluster, it only needs to compute the attention score once on the centroid vector. With a reduced number of vectors when performing self-attention, it reduces the complexity of self-attention from quadratic to linear. However, the cost is a noticeable error by approximating many queries with the same cluster, thereby leading to performance degradation. 
On the other hand, \citet{LinFormer_arXiv_2020} project key and value matrices into a low-dimensional space. Specifically, with $r$ keys and values in the low-dimensional space, the method only needs to perform an attention op on $r$ keys rather than $L$ keys as in the standard self-attention mechanism. However, due to the fact that the projection matrix is pre-defined and learned from scratch, it is not guaranteed that the low-dimensional projection is effective in preserving information in the original key and value matrices \cite{LinFormer_arXiv_2020}. 
Besides query/key/value matrices, \citet{FMMFormer_NIPS_2021, HyperAttention} directly exploit redundancy in the attention map, and approximate the attention map with low-rank and sparse matrices. Therefore, computation and memory costs of self-attention can also be reduced. 

Besides removing redundancy in self-attention operations, current works along the second line of research attack the problem via approximating \texttt{Softmax} operations with kernelization. Typically, \citet{Performers_ICLR_2021} regard self-attention as \texttt{Softmax} kernels: $\texttt{exp}(\bm{q} \cdot \bm{k}^T)$ with query $\bm{q}$ and key $\bm{k}$, and approximate it with the Gaussian kernel function\cite{RandomFeature_NIPS_2007}. 
Specifically, it estimates \texttt{Softmax} as: $\texttt{exp}(\bm{q}\cdot \bm{k}^T) \to \mathbb{E} \left [ \phi(\bm{q}) \cdot \phi(\bm{k})^T \right ]$, where kernel function $\phi(\cdot)$ maps a vector to a low-dimensional space. 
Therefore, the dimension after kernelization is reduced, leading to a reduction in self-attention operations. 
Along this line, other works \citep{TransformereRNN_PMLR_2020, FMMFormer_NIPS_2021} explore different kernel functions to approximate the self-attention function. 
While the complexity is reduced, these kernel-based approximations still incur large \texttt{Softmax} approximation errors given large hidden dimensions in large models.  

Therefore, \emph{a lightweight self-attention mechanism with linear complexity is still needed, especially for current large models with huge computation and memory footprints.}

\section{Lowrank Structure in Sequences}\label{sec:lowrank}
Low-rank structures in inputs of language models are an essential component, that, surprisingly, is rarely exploited in current models for better computation and memory efficiency. 
Compared to model parameters \cite{LoRA_ICLR_2021}, inputs and internal hidden states are usually more correlated, which can be potentially exploited. 
Such a property has also been observed in vision problems \cite{LegRace_PETS_2022, SAM_NIPS_2023} and used to reduce the complexity of convolution operations. 
This paper is the first work that investigates the low-rank structure of input sequences in language models, and, importantly, its potential to computation and memory saving. 
In this section, we first analyze low-rank structures in transformers' input sequence. Then, in the next section, we present \method{} that leverages low-rank structures in inputs and performs self-attention with significantly reduced computation and memory footprints. 

Transformer models comprise a stack of self-attention layers. Each self-attention layer takes input state $X \in \mathbb{R}^{L \times d}$, and computes output state $Y \in \mathbb{R}^{L \times d}$, where $L$ denotes the sequence length, $d$ is the dimension of each hidden state vector. Each state vector corresponds to a token in the input sequence. Owning to the semantic relationships among tokens, these vectors are also correlated. To formally measure such correlations, we adopt a metric called \texttt{SVD-Entropy}. 

In detail, we apply singular value decomposition (\texttt{SVD}) to the hidden state as

\begin{equation}\label{eq:svd}
    X  \quad \xrightarrow[]{\texttt{SVD}} \quad \sum_{i=1}^L \sigma_i \cdot \bm{u}_i \cdot \bm{v}_i^T.
\end{equation}

We assume $L\leq d$ without loss of generality. With Eq\eqref{eq:svd}, we attains singular values $\{\sigma_i\}$ and corresponding principal components $\{ \bm{v}_i \}$. Then, based on \citet{LegRace_PETS_2022}, we compute \texttt{SVD-Entropy} as the ``low-rankness" of $X$,

\begin{equation}\label{eq:entropy}
    \mu = -\log \left( \sum\limits_{i=1}^{L} \bar{\sigma}_i^2 \right ),
\end{equation}
where $\bar{\sigma}_i=\frac{\sigma_i}{\sum\limits_{i'=1}^L\sigma_{i'}}$.

According to \citet{LegRace_PETS_2022}, $\left \lceil 2^{\mu} \right \rceil$ can denote the number of necessary principal components to sufficiently approximate input $X$. $\left \lceil 2^{\mu} \right \rceil \ll L$ implies that input state vectors in $X$ are highly correlated such that only a few principal components are sufficient to represent $X$.
 
With such a measure, we analyze the low-rank structure of hidden states in language models. Figure \ref{fig:lr} shows the distribution of low-rankness after Llama-2's embedding layer on BoolQ and MMLU datasets, measured by ratio $\left \lceil 2^{\mu} \right \rceil / L$. A small ratio implies that the embedding of a sequence is more low-rank.  
We can easily observe that embeddings of all sequences are highly low-rank, where $50\%$ or even fewer principal components are sufficient to approximate embedding vectors without error. 
Moreover, longer sequences usually exhibit more low-rank structure compared to shorter sequences. 
Note that the observation implies that exploiting the low-rankness of input data can be more effective compared to the low-rankness of models \cite{LoRA_ICLR_2021}. 
Such a crucial observation presents great potential for reducing the dimension of inputs, thereby leading to more efficient self-attention with reduced computation and memory complexities, especially for long sequences.
Low-rankness analysis of other models is deferred to Appendix \ref{sec:appx:lowrank}.



\pgfplotstableread[col sep=comma]{datammlu1.csv}{\loadedtableA}
\pgfplotstableread[col sep=comma]{datammlu2.csv}{\loadedtableB}
\pgfplotstableread[col sep=comma]{datammlu3.csv}{\loadedtableC}
\pgfplotstableread[col sep=comma]{databoolq1.csv}{\loadedtableboolqA}
\pgfplotstableread[col sep=comma]{databoolq2.csv}{\loadedtableboolqB}
\pgfplotstableread[col sep=comma]{databoolq3.csv}{\loadedtableboolqC}

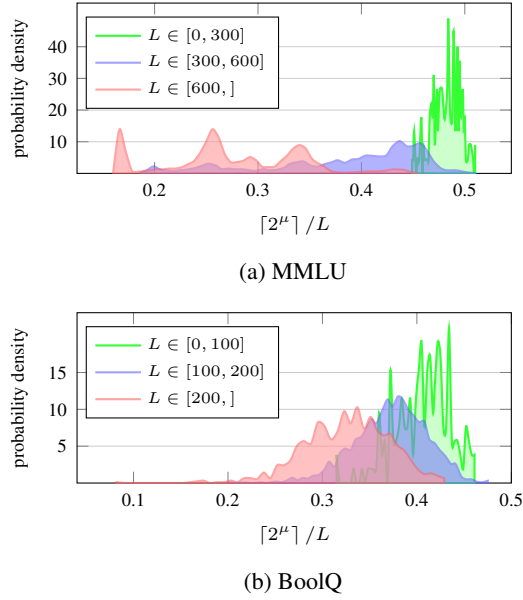
\begin{figure}[!htb]
    \centering
    \begin{subfigure}[b]{0.48\textwidth}
    \begin{tikzpicture}
        \begin{axis} [
            ticklabel style={font=\scriptsize},
            width=0.95\linewidth, height=.5\linewidth,
            xlabel={\scriptsize $\left \lceil 2^{\mu} \right \rceil / L$},
            xtick={ 0, 0.1, 0.2, 0.3, 0.4, 0.5 }, xticklabels = {0, 0.1, 0.2, 0.3, 0.4, 0.5},
            ylabel={\scriptsize probability density}, ymin=0,
            ytick={ 10,20, 30, 40 }, yticklabels={ 10,20, 30, 40 },
            ymajorgrids, major grid style={line width=.2pt, draw=black!20},
            legend style={at={ (0.02,0.65)}, anchor=west },
            legend cell align={left}
        ]
            \addplot[thick, fill=green!20, draw=green!80, smooth] table [x=bin, y=cnt, col sep=comma] {\loadedtableA} \closedcycle;
            \addlegendentry{\tiny $L\in [0, 300]$}

            \addplot[thick, fill=blue!50, draw=blue!60, smooth, opacity=0.6] table [x=bin, y=cnt, col sep=comma] {\loadedtableB} \closedcycle;
            \addlegendentry{\tiny $L\in [300, 600]$}

            \addplot[thick, fill=red!40, draw=red!60, smooth, opacity=0.6] table [x=bin, y=cnt, col sep=comma] {\loadedtableC} \closedcycle;
            \addlegendentry{\tiny $L\in [600,]$}

        \end{axis}
    \end{tikzpicture}   
    \caption{MMLU}
    \label{fig:lr:MMLU}
    \end{subfigure}
    
    \vspace{3mm}
    
    \begin{subfigure}[b]{0.48\textwidth}
    \begin{tikzpicture}
        \begin{axis} [
            ticklabel style={font=\scriptsize},
            width=0.95\linewidth, height=.5\linewidth,
            xlabel={\scriptsize $\left \lceil 2^{\mu} \right \rceil / L$},
            xmax=0.5,
            xtick={ 0, 0.1, 0.2, 0.3, 0.4, 0.5 }, xticklabels = {0, 0.1, 0.2, 0.3, 0.4, 0.5},
            ylabel={\scriptsize probability density}, ymin=0,
            ytick={ 5,10,15 }, yticklabels={ 5,10,15 },
            ymajorgrids, major grid style={line width=.2pt, draw=black!20},
            legend style={at={ (0.02,0.65)}, anchor=west },
            legend cell align={left}
        ]
            \addplot[thick, fill=green!20, draw=green!80, smooth] table [x=bin, y=cnt, col sep=comma] {\loadedtableboolqA} \closedcycle;
            \addlegendentry{\tiny $L\in [0, 100]$}

            \addplot[thick, fill=blue!50, draw=blue!60, smooth, opacity=0.6] table [x=bin, y=cnt, col sep=comma] {\loadedtableboolqB} \closedcycle;
            \addlegendentry{\tiny $L\in [100, 200]$}

            \addplot[thick, fill=red!40, draw=red!60, smooth, opacity=0.6] table [x=bin, y=cnt, col sep=comma] {\loadedtableboolqC} \closedcycle;
            \addlegendentry{\tiny $L\in [200, ]$}

        \end{axis}
    \end{tikzpicture}   
    \caption{BoolQ}
    \label{fig:lr:boolq}
    \end{subfigure}

    \caption{Distribution of low-rankness of Llama-2's embedding on MMLU and BoolQ dataset, measured by ratio $\left \lceil 2^{\mu} \right \rceil / L$. Almost all sequences can be sufficiently approximated with less than half principal components without incurring error. Longer sequences exhibit a more low-rank structure. }
    \label{fig:lr}
\end{figure}

\section{\method{} Methodology}\label{sec:method}
In this section, we introduce \method{}, a generic transformer architecture with a new efficient self-attention. 
\method{} introduces a rank-aware self-attention mechanism that reduces the complexity of self-attention to linear given the low-rank structure in input sequence embeddings.

\subsection{Self-Attention with Low-Rank Inputs}\label{subsec:selfattlr}
Given low-rank input $X \in \mathbb{R}^{L \times d}$ with $r$ principal components, we write it as 
\begin{equation}\label{eq:Xlr}
    X = U \cdot X',
\end{equation}
where $U\in \mathbb{R}^{L \times r}$, and $X' \in \mathbb{R}^{r \times d}$ denotes the principal components. 
Since $X$ is low-rank, query/key/values matrices obtained by projecting $X$ are also low-rank. That is,

\begin{equation}\label{eq:QKVlr}
\begin{split}
    Q,K,V &= U \cdot X' \cdot \left \{ W^Q, W^K, W^V  \right \} \\
    &=U \cdot \left \{ Q', K', V'  \right \}.
\end{split}
\end{equation}
By the matrix rank inequality \cite{MatrixRank}, we have $\texttt{rank}(\left \{ Q, K, V  \right \}) \leq \texttt{rank}(X') = r$.

Then we start from the standard self-attention, and show the computations can be significantly reduced with low-rank keys/values. 
We omit the normalization in \texttt{Softmax} and write self-attention with query $\bm{q}$ on all keys/values as $\texttt{exp}(\bm{q}, K^T) \cdot V$.
With low-rankness of input $X$, we can break down the self-attention as 

\begin{equation}\label{eq:SAlr}
\begin{split}
    \texttt{exp}(\bm{q} \cdot K^T) \cdot V = \texttt{exp}(\bm{q} \cdot K'^T \cdot U^T) \cdot U \cdot V'
\end{split}
\end{equation}

By the Taylor expansion on the \texttt{exp} function on ech value, we have the following approximation,

\begin{equation}\label{eq:Taylor}
\begin{split}
    &\texttt{exp}(\bm{q}, K^T) \cdot V \\
    & \simeq \bm{1} \cdot U \cdot V' + \bm{q} \cdot K'^T \cdot U^T \cdot U \cdot V' \\
    & = \bm{1} \cdot U \cdot V' + \bm{q} \cdot K'^T \cdot V' \\
    & = (\bm{1} \cdot U + \bm{q} \cdot K'^T) \cdot V' = A' \cdot V',
\end{split}
\end{equation}
where $\bm{1} \in \mathbb{R}^{1\times L}$, and $ U^T \cdot U = I$. 
Similar as \texttt{Softmax}, normalization is applied row-wise on the new attention map $A'$. 

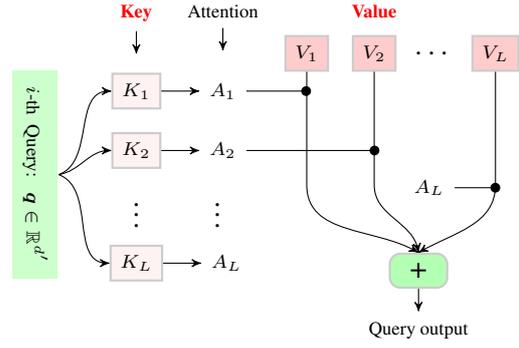
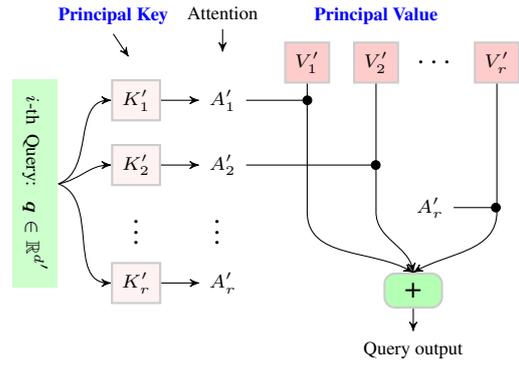
\begin{figure}[!htb]
    \centering
    \begin{subfigure}[b]{.48\textwidth}
    \centering
    \begin{tikzpicture}
        \node[data, draw=none, fill=green!20, text width=25mm, rotate=270] (query) {\scriptsize $i$-th Query: $\bm{q}\in\mathbb{R}^{d'}$};
        
        \node[data,draw=black!20, fill=pink!20, right=10mm of query, yshift=25mm] (key1) {\scriptsize $K_1$};
        \node[data,draw=black!20, fill=pink!20, below=3mm of key1] (key2) {\scriptsize $K_2$};
        \node[below=1mm of key2] (keyi) {$\vdots$};
        \node[data,draw=black!20, fill=pink!20, below=1mm of keyi] (keym) {\scriptsize $K_L$};
        \node[above=5.5mm of key1](Key){\scriptsize \red{\textbf{Key}}};
        \draw[->] (Key.south) -- +(0mm, -3mm);
        
        \node[right=5mm of key1] (att1) {\scriptsize $A_1$};
        \node[right=5mm of key2] (att2) {\scriptsize $A_2$};
        \node[right=7mm of keyi] (atti) {$\vdots$};
        \node[right=5mm of keym] (attm) {\scriptsize $A_L$};
        \node[right=20mm of attm, yshift=10mm] (attmm) {\scriptsize $A_L$};
        \node[above=6mm of att1](Attention){\scriptsize Attention};
        \draw[->] (Attention.south) -- +(0mm, -3mm);

        \node[data,draw=black!20, fill=red!20, right=5mm of att1, yshift=5mm] (value1) {\scriptsize $V_1$};
        \node[data,draw=black!20, fill=red!20, right=3mm of value1] (value2) {\scriptsize $V_2$};
        \node[right=1mm of value2] (valuei) {$\cdots$};
        \node[data,draw=black!20, fill=red!20, right=1mm of valuei] (valuem) {\scriptsize $V_L$};
        \node[above=1mm of value2](){\scriptsize \red{\textbf{Value}}};

        \draw[->] (query.north) to [out=20, in=180] (key1.west);
        \draw[->] (query.north) to [out=10, in=180] (key2.west);
        \draw[->] (query.north) to [out=-20, in=180] (keym.west);

        \draw[->] (key1) -- (att1);
        \draw[->] (key2) -- (att2);
        \draw[->] (keym) -- (attm);

        \node[op, draw=black!20, fill=green!30, right=18mm of attm, yshift=-1mm, text width=5mm] (add) {+};
        \node[below=3mm of add] (output) {\scriptsize Query output};
        \draw[->] (value1.south) -- +(0mm, -15mm) to [out=-90, in=150] (add.north);
        \draw[->] (value2.south) -- +(0mm, -15mm) to [out=-90, in=100] (add.north);
        \draw[->] (valuem.south) -- +(0mm, -15mm) to [out=-90, in=30] (add.north);
        \draw[-Circle] (att1.east) -- +(8.5mm, 0mm);
        \draw[-Circle] (att2.east) -- +(17.5mm, 0mm);
        \draw[-Circle] (attmm.east) -- +(6mm, 0mm);
        \draw[->] (add.south) -- (output.north);
        
    \end{tikzpicture}
    \caption{Standard self-attention.}
    \label{fig:att:standard}
    \end{subfigure}

    \vspace{3mm}

    \begin{subfigure}[b]{.48\textwidth}
    \centering
    \begin{tikzpicture}
        \node[data, draw=none, fill=green!20, text width=25mm, rotate=270] (query) {\scriptsize $i$-th Query: $\bm{q}\in\mathbb{R}^{d'}$};
        
        \node[data,draw=black!20, fill=pink!20, right=10mm of query, yshift=25mm] (key1) {\scriptsize $K'_1$};
        \node[data,draw=black!20, fill=pink!20, below=3mm of key1] (key2) {\scriptsize $K'_2$};
        \node[below=1mm of key2] (keyi) {$\vdots$};
        \node[data,draw=black!20, fill=pink!20, below=1mm of keyi] (keym) {\scriptsize $K'_r$};
        \node[above=6mm of key1, xshift=-3mm, text centered](Key){\scriptsize \blue{\textbf{Principal Key}}};
        \draw[->] (Key.south) -- +(2mm, -3mm);
        
        \node[right=5mm of key1] (att1) {\scriptsize $A'_1$};
        \node[right=5mm of key2] (att2) {\scriptsize $A'_2$};
        \node[right=7mm of keyi] (atti) {$\vdots$};
        \node[right=5mm of keym] (attm) {\scriptsize $A'_r$};
        \node[right=21mm of attm, yshift=10mm] (attmm) {\scriptsize $A'_r$};
        \node[above=6.7mm of att1](Attention){\scriptsize Attention};
        \draw[->] (Attention.south) -- +(0mm, -3mm);

        \node[data,draw=black!20, fill=red!20, right=5mm of att1, yshift=5mm] (value1) {\scriptsize $V'_1$};
        \node[data,draw=black!20, fill=red!20, right=3mm of value1] (value2) {\scriptsize $V'_2$};
        \node[right=1mm of value2] (valuei) {$\cdots$};
        \node[data,draw=black!20, fill=red!20, right=1mm of valuei] (valuem) {\scriptsize $V'_r$};
        \node[above=1mm of value2](){\scriptsize \blue{\textbf{Principal Value}}};

        \draw[->] (query.north) to [out=20, in=180] (key1.west);
        \draw[->] (query.north) to [out=10, in=180] (key2.west);
        \draw[->] (query.north) to [out=-20, in=180] (keym.west);

        \draw[->] (key1) -- (att1);
        \draw[->] (key2) -- (att2);
        \draw[->] (keym) -- (attm);

        \node[op, draw=black!20, fill=green!30, right=18mm of attm, yshift=-1mm, text width=5mm] (add) {+};
        \node[below=3mm of add] (output) {\scriptsize Query output};
        \draw[->] (value1.south) -- +(0mm, -17mm) to [out=-90, in=150] (add.north);
        \draw[->] (value2.south) -- +(0mm, -17mm) to [out=-90, in=100] (add.north);
        \draw[->] (valuem.south) -- +(0mm, -17mm) to [out=-90, in=30] (add.north);
        \draw[-Circle] (att1.east) -- +(8.7mm, 0mm);
        \draw[-Circle] (att2.east) -- +(17.7mm, 0mm);
        \draw[-Circle] (attmm.east) -- +(6.2mm, 0mm);
        \draw[->] (add.south) -- (output.north);
        
    \end{tikzpicture}
    \caption{Low-rank self-attention.}
    \label{fig:att:lr}
    \end{subfigure}
    \caption{Standard self-attention and low-rank self-attention. Low-rank self-attention share the same procedure as the standard self-attention, but with only $r$ principal keys and values.}
    \label{fig:att}
\end{figure}

Eq\eqref{eq:Taylor} shows that self-attention on all token vectors $X$ can be converted to attention on all principal keys $K'$. More importantly, different from the standard self-attention where each key corresponds to a token in the input sequence, these principal keys denote all principal bases drawn from $X'$.  That is, \emph{\method{} converts the attention operation from individual token vectors to principal basis vectors.}
The observation is very crucial given low-rank input $X$. The reason is that, given low-rank input with $r \ll L$, based on Eq\eqref{eq:Taylor}, for each query, we only need to perform dot-product on $r$ vectors, rather than $L$ vectors as in the standard self-attention. 
Therefore, the self-attention does not incur $\mathcal{O}(L^2)$ computation and memory costs. Instead, the costs scale linearly with sequence length, $L$, and the number of principal components, $r$.
Figure \ref{fig:att} shows a point-to-point comparison between the standard self-attention and the low-rank self-attention. The low-rank self-attention shares a similar procedure as the stand self-attention, while the difference is that the low-rank self-attention performs dot-product on $r$ principal keys.

\begin{remark}
    Unlike works such as \cite{LinFormer_arXiv_2020} that attain low-dimensional key/value matrices via hard-coded/learnable projection, we adopt a more rigorous method based on \texttt{SVD} to find the optimization low-dimensional space, that preserves most energy of input $X$ with $r$ principal components. 
\end{remark}

\subsection{Tansformers with Low-Rank Attention}\label{subsec:transformerlr}
With the low-rank self-attention above, we can adapt the transformer architecture to input sequences with highly low-rank structure.  To the best of our knowledge, this is the first adaptation that takes input low-rank structure into model design, and reduces complexities for the whole pipeline. 

As a transformer model is usually built with a stack of encoder/decoder layers with the same architecture, to simplify, we only show the architecture adaptation for one encoder/decoder layer, which will be replicated to the rest of the layers. 

The first step is to analyze input $X$ and attain its principal components. 
To that end, we decompse input $X$ using \texttt{SVD}, and attain the principal components $X'$ based on Eq\eqref{eq:svd}: $X' = \left [ \sigma_1\bm{v}_1, \cdots, \sigma_r\bm{v}_r \right ]$.
However, the exact SVD incurs a complexity of $\mathcal{O}(Ld^2)$, which can be a performance bottleneck given the large dimension of each vector in $X$.
To avoid such a quadratic complexity, we adopt an approximated \texttt{SVD} algorithm as 

\begin{equation}\label{eq:svdapprox}
\begin{split}
    X' &= \left [ \sigma_1\bm{v}_1, \cdots, \sigma_r\bm{v}_r \right ] \\
    &= \texttt{argmin}_{\sigma,\bm{u,v}_{1,\cdot, r}} \left \| X - \sum_{i=1}^r \sigma_i \cdot \bm{u}_i \cdot \bm{v}_i^T \right \|.
\end{split}
\end{equation}

Essentially, the optimization above is to find $r$ principal components that preserve most energy in $X$, while ignoring the orthogonality constraint on the components. 
To simplify the optimization above, $\sigma_i$ can be fused with $\bm{v}_i$ to reduce the number of variables to be optimized. By the alternating optimization in Alg 1 in \citet{LegRace_PETS_2022} (duplicated in Appendix \ref{sec:appx:alter}), we can attain $r$ most principal components which preserved most energy in $X$.
Compared to the standard SVD decomposition, the approximation in Eq\eqref{eq:svdapprox} incurs a linear complexity of $\mathcal{O}(rLd)$, thereby preventing \texttt{SVD} being the bottleneck in the whole pipeline.

\begin{figure}
    \centering
    \begin{tikzpicture}
        \node[op, fill=green!10, text width=25mm] (lrSA) {\scriptsize Low-Rank Attention};
        \node[op, fill=pink!30, below=3mm of lrSA, xshift=-15mm, text width=5mm] (Q) {\scriptsize $W^Q$};
        \node[op, fill=pink!30, below=3mm of lrSA, xshift=0mm, text width=5mm] (K) {\scriptsize $W^K$};
        \node[op, fill=pink!30, below=3mm of lrSA, xshift=15mm, text width=5mm] (V) {\scriptsize $W^V$};
        \node[op, fill=pink!30, below=8mm of K, text width=5mm] (svd1) {\scriptsize \texttt{SVD}};
        \node[op, fill=pink!30, above=5mm of lrSA, text width=5mm] (NORM) {\scriptsize Norm};
        \node[op, fill=pink!30, above=2mm of NORM, text width=15mm] (FF) {\scriptsize Feedforward};
        \node[below=3mm of svd1, opacity=1.0] (X) {\scriptsize $X$};
        \node[above=5mm of FF, opacity=1.0] (Y) {\scriptsize $Y$};
        \node[above=0.3mm of svd1, xshift=-3mm](){\scriptsize $X'$};
        \node[left=1mm of Q]() {\scriptsize Projection};
        \draw[->] (X) -> (svd1);
        \draw[->] (svd1) -- +(0mm, 6mm) -> (Q);
        \draw[->] (svd1) -> (K);
        \draw[->] (svd1) -- +(0mm, 6mm) -> (V);
        \draw[->] (Q) -> (lrSA);
        \draw[->] (K) -> (lrSA);
        \draw[->] (V) -> (lrSA);
        \draw[->] (lrSA) -> (NORM);
        \draw[->] (NORM) -> (FF);
        \draw[->] (FF) -> (Y);
        \draw[->] (svd1.north) -- +(0mm, 2mm) -- +(20mm, 2mm) -- +(20mm, 22mm) -- +(0mm, 22mm);
        \draw[->] (lrSA.north) -- +(0mm, 3mm) -- +(10mm, 3mm) -- +(10mm, 18mm) -- +(0mm, 18mm);

    \end{tikzpicture}
    \caption{Transformer encoder/decoder with low-rank self-attention. Input $X$ is first fed to \texttt{SVD} to attain the principal components, $X'$. Then, $X'$ is fed to an encoder/decoder layer with low-rank self-attention.}
    \label{fig:transformer:lr}
\end{figure}
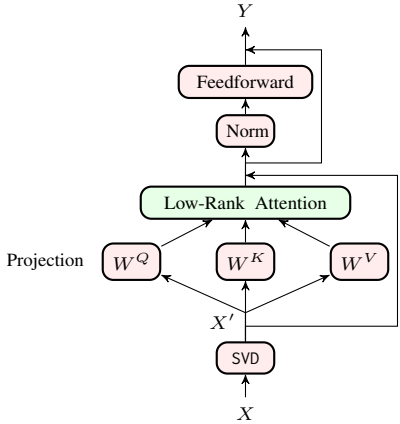

Then, principal components $X'$ are fed into a self-attention layer to attain principal keys and values as in Eq\eqref{eq:QKVlr}. With the principal keys and values, \method{} performs attention as in Eq\eqref{eq:Taylor}, and feedforward to obtain output states $Y$. The next encoder/decoder layer follows the same procedure first to attain principal components of $Y$ and perform low-rank self-attention. 

\textbf{Combine with Position Encoding}. For absolution or relative position encoding vectors $P$ for a sequence \cite{BERT_2018_arXiv, ALBERT_2020_ICLR, RelativePos_2018_NACCL}, they are added to token embeddings before an encoder/decoder layer. Therefore, we can still directly apply \texttt{SVD} to the input vectors, $X+P$, and obtain principal components for low-rank self-attention. 

For rotatory position embedding \cite{RoFormer_2021_arXiv, Llama_2023_arXiv}, the position encoding vectors are added after query/key projection. That is, $\bm{k} = \bm{x} \cdot W^K \cdot R_i$, where $R_i$ is a rotary matrix corresponding to a transformation for a token at position $i$, $\bm{x}$ denotes one input vector in $X$. 
While the low-rank structure might change during the rotation, we can still attain a low-rank key matrix by projecting the key matrix into a low-dimension space with $U$ as in Eq\eqref{eq:Taylor}.

Therefore, the low-rank self-attention mechanism is compatible with current position encoding methods. 

\begin{table*}[!htb]
    \centering
    \begin{tabular}{c|ccc|ccc}
    \toprule
     Mechanism  & \multicolumn{3}{c|}{Standard} & \multicolumn{3}{c}{Low-rank} \\ \midrule
         & Computation & Complexity & Memory & Computation & Complexity & Memory\\
      Projection   & $X\cdot W$ & $\mathcal{O}(Ldd')$ & $\mathcal{O}(Ld')$ & $X' \cdot W$ & $\mathcal{O}(rdd')$ & $\mathcal{O}(rd')$\\
      Attention    & $Q\cdot K^T $ & $\mathcal{O}(L^2d')$ & $\mathcal{O}(L^2)$ & $QK'^T$ & $\mathcal{O}(rLd')$ & $\mathcal{O}(rL)$\\
    \bottomrule
    \end{tabular}
    \caption{Computation and memory complexity with low-rank input. Low-rank self-attention reduces the complexity of attention from quadratic to linear. It also reduces complexities for other linear layers ($L$: sequence length, $r$: rank, $d$: dimension of $X$, $d'$: dimension of hidden state).}
    \label{tab:complexity}
\end{table*}

\subsection{Complexity Analysis}\label{subsec:complex}
Self-attention with low-rank inputs not only relieves computation and memory pressure for attention operations, but also reduces complexity for other linear layers. 
Table \ref{tab:complexity} lists computations and the corresponding complexity of the standard and low-rank self-attention. 
Due to the reduced number of components in $X'$, query/key/value projection only needs to project $r$ vectors rather than $L$ token vectors as the standard self-attention, thereby resulting in $r$ keys and value vectors with dimension $d'$. Hence, both the computation and memory during the projection are reduced by $L/r$.
On the other hand, when performing attention, the low-rank attention only needs to compute the attention score on $r$ principal keys, rather than $L$ token keys as the standard self-attention. Therefore, the computation and memory complexities are also reduced by $L/r$.
Note that the additional \texttt{SVD} only incurs computation complexity of $\mathcal{O}(rLd)$, which is linear in term of $L$, and is relatively small compared to computations in the standard self-attention.
In addition, more computation saving can be achieved by decomposing hidden state vectors to the \texttt{FeedForward} layer. 
In this paper, we mainly focus on self-attention layers. 

\begin{figure}[!htb]
    \centering
    \begin{tikzpicture}
    \begin{axis} [
            ticklabel style={font=\scriptsize},
            width=0.95\linewidth, height=.6\linewidth,
            xlabel={\scriptsize sequence length},
            xtick={ 1, 2, 3, 4, 5 }, xticklabels = { 512, 1024, 2048, 4096, 8192 },
            ylabel={\scriptsize normalized time}, ymin=0,
            ytick={ 2, 4, 6, 8 }, yticklabels={ .2, .4, .6, .8 },
            ymajorgrids, major grid style={line width=.2pt, draw=black!20},
            legend style={at={ (0.02,0.65)}, anchor=west },
            legend cell align={left}
        ]

        \addplot+[ 
                thick, blue, smooth, mark=*, mark options={solid, scale=0.5},
                error bars/.cd, y fixed, y dir=both, y explicit
            ] 
            table [x=x, y=y, y error plus=error1, y error minus=error2, col sep=comma]{
                x,      y,          error1,     error2
                1,   0.23,       0.1,        0.05
                2,   0.46,       0.1,        0.05
                3,   1.03,       0.1,        0.1
                4,   2.11,       0.3,        0.2
                5,   3.51,       0.5,        0.3
            };
            \addlegendentry{\tiny low-rank}

        \addplot+[ 
                thick, red, smooth, mark=*, mark options={solid, scale=0.5},
                error bars/.cd, y fixed, y dir=both, y explicit
            ] 
            table [x=x, y=y, y error plus=error1, y error minus=error2, col sep=comma]{
                x,      y,          error1,     error2
                1,   0.29,       0.1,        0.1
                2,   0.67,       0.1,        0.1
                3,   1.72,       0.2,        0.1
                4,   4.21,       0.4,        0.6
                5,   10.87,       0.5,        0.6
            };
            \addlegendentry{\tiny standard}
            
    \end{axis}
    \end{tikzpicture}
    \caption{Actual running time of low-rank self-attention compared to the standard mechanism with different sequence lengths ($r$=128). The running time of the standard self-attention increases quadratically with the sequence length. Low-rank self-attention reduces the running time to almost linear. }
    \label{fig:complexity}
\end{figure}
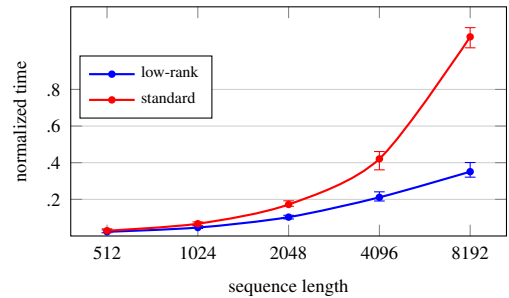

Figure \ref{fig:complexity} shows actual speedups of low-rank self-attention compared to the standard self-attention given different sequence lengths. Note that the standard self-attention, as expected, incurs quadratic running time with increasing input sequence length. On the other hand, the running time of the low-rank self-attention scales almost linearly with sequence length. 
The time gap between them grows rapidly with long sequences. 
This shows that the standard self-attention indeed comes with a severe bottleneck on real performance with long sequences,  while the low-rank self-attention significantly reduces actual running time. 

\section{Empirical Evaluation}\label{sec:eval}
In this section, we evaluate the low-rank attention on benchmark models and datasets. 
To investigate the applicability of low-rank attention in a wide range of applications, we choose models with different sizes. 
For datasets, we focus on long sequences, which usually incur significant computation and memory pressure during inference. 

\textbf{Model}. We choose BERT-base (encoders only) as the small model \cite{BERT_2018_arXiv}, Llama2-7B (decoder only) as the medium model, and Llama2-13B as the large model \cite{Llama_2023_arXiv}. 
Table \ref{tab:model:arch} lists their detailed architecture parameters.
Note that all three models adopt the standard self-attention mechanism. 

\begin{table}[!htb]
    \centering
    \small
    \begin{tabular}{c|c|c|c}
    \toprule
        & BERT & Llama2-7B & Llama2-13B  \\ \midrule
        \# att layers & 12 & 32 & 40 \\
        \# heads/layer & 12 & 32 & 40  \\
        \# head dim & 64 & 128 & 128 \\
    \bottomrule
    \end{tabular}
    \caption{Architecture parameters of BERT-base, Llama2-7b and Llama2-13B.}
    \label{tab:model:arch}
\end{table}

\textbf{Datasets}. For BERT-base, we choose SST-2, Squad \cite{Superglue_NIPS_2019}, and IMDB\cite{IMDB_ACL_2011}. In particular, the IMDB dataset consists of long sequences that exhibit more low-rank structures. For Llama2-7B and Llama2-13B, we choose two of the official benchmark datasets: MMLU \cite{MMLU_ICLR_2021} and BoolQ \cite{BoolQ_NAACL_2019}.

\subsection{BERT-base}\label{sec:val:bert}
For all datasets, we start from a pre-trained model, replace each self-attention layer with the low-rank self-attention, and finetune the model. Owing to the model size, we finetune full parameters. Training details are provided in Appendix \ref{sec:appx:hparam}.
Table \ref{tab:acc:bert} lists the final model accuracy on SST-2, Squad, and IMDB. We can observe that BERT-base with low-rank self-attention preserves models' performance. In particular, with $1/2$ principal keys used, the model with low-rank self-attention barely loses accuracy. This indicates that owing to the low-rank structure in sequences, $1/2$ principal keys preserve most information in inputs. Surprisingly, we can further see that even only keeping $1/8$ principal keys, the model still achieves a comparable accuracy as the model with standard self-attention. 

\begin{table*}[!htb]
    \centering
    \begin{tabular}{c|cccc}
    \toprule
       Model  & Original & 1/2 & 1/4 & 1/8 \\ \midrule
       SST-2  & $92.32 \pm 0.2$ & $92.1 \pm 0.17$ & $91.0 \pm 0.23$ & $89.2 \pm 0.26$ \\
       Squad & $88.15 \pm 0.3$ & $87.93 \pm 0.2$ & $87.23 \pm 0.34$ & $84.94 \pm 0.28$ \\
       IMDB & $91.45 \pm 0.2$ & $90.97 \pm 0.19$ & $89.65 \pm 0.3$ & $87.28 \pm 0.3$\\
    \bottomrule
    \end{tabular}
    \caption{BERT-base accuracy on SST-2, Squad, and IMDB using low-rank self-attention.}
    \label{tab:acc:bert}
\end{table*}

\begin{figure}[!htb]
    \centering
    \begin{tikzpicture}
        \begin{axis}[
            width=.9\linewidth, height=.6\linewidth,
            xlabel={\scriptsize complexity},
            xtick={0.125, 0.25, 0.5, 0.75, 1}, xticklabels={1/8, 1/4, 1/2, 3/4, full },
            ymin=0.4, ymax=1.03,
            ylabel={\scriptsize energy ratio(\%)},
            ytick={0.8, 0.9, 1}, yticklabels={80, 90, 100},
            ymajorgrids,
            major grid style={line width=.2pt, draw=black!30},
            extra y ticks={0.6}, extra y tick labels={60},
            extra tick style={major grid style=red!50, dashed},
            ticklabel style={font=\tiny},
            legend style={at={ (1.0,0.2)}, anchor=east }
        ]
    
            \addplot+[ 
                thick, red, smooth,
                mark=*, mark options={solid, scale=0.75},
                error bars/.cd,
                y fixed,
                y dir=both,
                y explicit
            ] 
            table [x=x, y=y, y error=error, col sep=comma]{
                x,      y,      error
                0.125,      0.873,   0.0
                0.25,      0.951,   0.0
                0.5,      0.996,   0.0
                1,      1,     0.0
            };
            \addlegendentry{\tiny SST-2}
            
            \addplot+[ 
                thick, blue, smooth, mark=*, mark options={solid, scale=0.75},
                error bars/.cd, y fixed, y dir=both, y explicit
            ] 
            table [x=x, y=y, y error=error, col sep=comma]{
                x,      y,       error
                0.125,  0.609,   0.0
                0.25,   0.780,   0.0
                0.5,    0.948,   0.0
                1,      1,       0.0
            };
            \addlegendentry{\tiny Squad}

            \addplot+[ 
                thick, green, smooth, mark=*, mark options={solid, scale=0.75},
                error bars/.cd, y fixed, y dir=both, y explicit
            ] 
            table [x=x, y=y, y error=error, col sep=comma]{
                x,      y,       error
                0.125,  0.81,   0.0
                0.25,   0.87,   0.0
                0.5,    0.963,   0.0
                1,      1,       0.0
            };
            \addlegendentry{\tiny IMDB}
            
        \end{axis}
        \end{tikzpicture}
    \caption{Energy ratio ($\left \| X' \right \|^2_F / \left \| X \right \|^2_F$) in low-rank hidden representations. Embeddings of all three datasets exhibit highly low-rank structures, with $1/2$ principal components preserving almost all energy. }
    \label{fig:energy}
\end{figure}
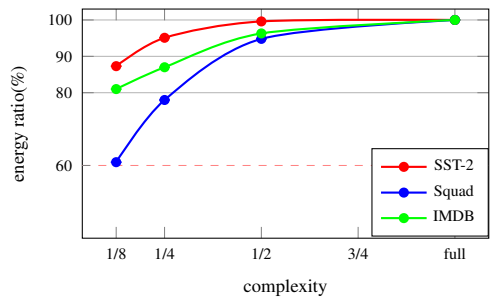

Figure \ref{fig:energy} shows the relative energy kept in the low-rank keys. We observe that for $1/2$ principal keys are sufficient to keep almost all energy in inputs, which is aligned with model accuracy in Table \ref{tab:acc:bert}. 
On the other hand, compared to Squad and IMDB, SST-2 exhibits a more low-rank structure, with even $1/8$ principal keys still preserving near $90\%$ energy. The observation explains BERT-base's performance on SST-2 that even low-rank self-attention with only $1/8$ principal keys only incurs a $\sim 3\%$ accuracy drop. 

\subsection{Llama2}\label{sec:val:llama}
We obtain pre-trained Llama2-7B/13B models from the Hugging Face repo \footnote{\url{https://huggingface.co/meta-llama}}. 
Starting from the pre-trained models, we replace their attention layers with low-rank self-attention. 
For MMLU and BoolQ, since they have different formats, we will first finetune the model on the datasets for a few iterations (See Appendix \ref{sec:appx:hparam} for more finetuning parameters), and then evaluate their performance on the validation dataset. Appendix \ref{sec:appx:prompt} provides prompt formats for MMLU and BoolQ during training and validation.
To reduce training workload, we use LoRA \cite{LoRA_ICLR_2021} to finetune the projection matrix for queries/keys/values with rank of $32$, and fix other layers.

For MMLU, we obtain the first predicted logit vector from the model given an input sequence, and compute the probability on the four tokens: \emph{A, B, C, D}. The token with the highest probability will be the predicted answer. For BoolQ, we adopt a similar procedure but compute the probability on the two tokens: \emph{Yes, No}, and output the token with the highest probability. Note that we ignore other tokens that might have the highest probability. 

\begin{figure}[!htb]
    \centering
    \captionsetup[subfigure]{oneside,margin={1cm,0cm}}
    \begin{subfigure}[b]{0.48\textwidth}
        \centering
        \begin{tikzpicture}
        \begin{axis} [
            ybar, bar width=3pt, ticklabel style={font=\scriptsize},
            width=.9\linewidth, height=.5\linewidth,
            xmin=0.5, xmax=4.5,
            xtick={ 1, 2, 3, 4 }, xticklabels={ STEM, humanities, social, other },
            ylabel={\scriptsize Acc}, ymin=0.2,
            ytick={ 0.3, 0.4, 0.5, 0.6 }, yticklabels={0.3, 0.4, 0.5, 0.6},
            ymajorgrids, major grid style={line width=.2pt, draw=black!20},
            extra y ticks={0.25}, extra y tick labels={0.25},
            extra tick style={major grid style=red!50, dashed},
            legend style={at={ (0.05,1.2)}, anchor=west, column sep=5pt },
            legend columns=-1
        ]
            \addplot[fill=blue!30, draw=blue!60] coordinates {
                (1, 0.37) 
                (2, 0.434) 
                (3, 0.517) 
                (4, 0.525)
            };
            \addlegendentry{\tiny Orig}

            \addplot[fill=magenta!30, draw=magenta!60] coordinates {
                (1, 0.36) 
                (2, 0.432) 
                (3, 0.523) 
                (4, 0.518)
            };
            \addlegendentry{\tiny 1/2}

            \addplot[fill=red!30, draw=red!60] coordinates {
                (1, 0.357) 
                (2, 0.422) 
                (3, 0.514) 
                (4, 0.507)
            };
            \addlegendentry{\tiny 1/3}

            \addplot[fill=green!30, draw=green!60] coordinates {
                (1, 0.34) 
                (2, 0.417) 
                (3, 0.511) 
                (4, 0.505)
            };
            \addlegendentry{\tiny 1/4}
        \end{axis}
        \end{tikzpicture}   
        \caption{Llama2-7B.}
        \label{fig:mmlu:llama7b}
    \end{subfigure}

    \vspace{3mm}

    \begin{subfigure}[b]{0.48\textwidth}
    \centering
    \begin{tikzpicture}
        \begin{axis} [
            ybar, bar width=3pt, ticklabel style={font=\scriptsize},
            width=.9\linewidth, height=.5\linewidth,
            xmin=0.5, xmax=4.5,
            xtick={ 1, 2, 3, 4 }, xticklabels={ STEM, humanities, social, other },
            ylabel={\scriptsize Acc}, ymin=0.2,
            ytick={ 0.3, 0.4, 0.5, 0.6 }, yticklabels={0.3, 0.4, 0.5, 0.6},
            ymajorgrids, major grid style={line width=.2pt, draw=black!20},
            extra y ticks={0.25}, extra y tick labels={0.25},
            extra tick style={major grid style=red!50, dashed},
            legend style={at={ (0.02,0.9)}, anchor=west }
        ]
            \addplot[fill=blue!30, draw=blue!60] coordinates {
                (1, 0.443) 
                (2, 0.544) 
                (3, 0.634) 
                (4, 0.608)
            };
            
            \addplot[fill=magenta!30, draw=magenta!60] coordinates {
                (1, 0.441) 
                (2, 0.545) 
                (3, 0.635) 
                (4, 0.606)
            };
            
            \addplot[fill=red!30, draw=red!60] coordinates {
                (1, 0.434) 
                (2, 0.535) 
                (3, 0.629) 
                (4, 0.601)
            };
            
            \addplot[fill=green!30, draw=green!60] coordinates {
                (1, 0.425) 
                (2, 0.529) 
                (3, 0.619) 
                (4, 0.594)
            };
        \end{axis}
    \end{tikzpicture}   
    \caption{Llama2-13B.}
    \label{fig:mmlu:llama13b}
    \end{subfigure}
    \caption{LLama2 on MMLU (random guess: 0.25). Low-rank self-attention effectively preserves performance on all subjects, even with $1/4$ principal keys. }
    \label{fig:mmlu}
\end{figure}
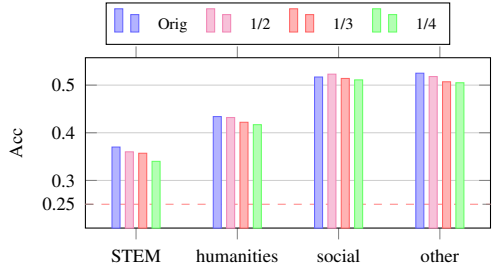
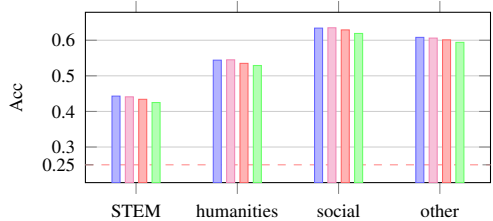

Figure \ref{fig:mmlu} shows the accuracy of Llama2-7B and 13B on MMLU using \method{}. We can observe that on all categories, \method{} achieves accuracy close to original Llama2-7B and 13B with standard self-attention. In particular, owing to the highly low-rank structure in input sequences, with $1/2$ principal keys, the model performance with \method{} is almost identical to the original model. Furthermore, even with only $1/4$ principal keys, \method{} still does not incur a significant accuracy drop. 
Similar performance of LLama2-7B and 13B with the low-rank self-attention holds on the BoolQ dataset, as listed in Table \ref{tab:boolq:llama}. 
Therefore, \method{} effectively leverages low-rank structure in input sequences and performs self-attention with a few top principal keys, leading to performance close to the original model but with significantly reduced complexities. 

\begin{table}[!htb]
    \centering
    \begin{tabular}{c|cccc}
    \toprule
       Model & Orig & 1/2 & 1/3 & 1/4 \\ \midrule
       7B & 0.795 & 0.791 & 0.789 & 0.763 \\
       13B & 0.839 & 0.836 & 0.819 & 0.816 \\
    \bottomrule
    \end{tabular}
    \caption{Llama2 on BoolQ with low-rank self-attention. Performance is not greatly affected even a small fraction of principal keys/values are used in attention layers.}
    \label{tab:boolq:llama}
\end{table}

\section{Conclusion}\label{sec:conclusion}
In this work, we propose a low-rank self-attention mechanism, \method{}, significantly reducing computation and memory complexity for transformers and LLMs. 
\method{} leverages low-rank structures in input sequences and sufficiently represents each input sequence with a few top principal components. 
Then, \method{} designs a low-rank self-attention layer that first attains principal keys/values given a low-rank input. Then, it performs attention only on top principal keys/values, rather than on each individual token embedding. Therefore, \method{} reduces the attention complexity from quadratic to linear in terms of sequence length. 
Owing to low-rank structures in input sequences, a few top principal keys/values are sufficient to preserve information in input sequences.
Evaluation of BERT and Llama models shows \method{} achieves performance close to original models with much-reduced computation and memory footprints. 

\section{Limitations}\label{sec:limitation}
\textbf{Limitations. } One of the limitations of this work is that we evaluate \method{} on BERT and Llama2 models. While performance on other models may differ. We will evaluate more models and datasets in future works.  

\noindent \textbf{Potential Risk. } While this work is aimed at lowering the barrier of deploying LLMs, it may be misused by malicious parties to quickly deploy and run adverse LLM services for their purposes. 

\bibliography{custom}

\appendix
\section{Lowrank Structure in Other Model}\label{sec:appx:lowrank}
Figure \ref{fig:lr:imdb} shows the low-rankness of BERT model on IMDB dataset. 
We can also observe that most sequences exhibists low-rank structures. In particular, long sequences are more low-rank, which is aligned with the observation in Sec \ref{sec:lowrank}. 


\pgfplotstableread[col sep=comma]{dataimdb1.csv}{\loadedtableA}
\pgfplotstableread[col sep=comma]{dataimdb2.csv}{\loadedtableB}
\pgfplotstableread[col sep=comma]{dataimdb3.csv}{\loadedtableC}

\begin{figure}[!htb]
    \begin{tikzpicture}
        \begin{axis} [
            ticklabel style={font=\scriptsize},
            width=0.95\linewidth, height=.6\linewidth,
            xlabel={\scriptsize $\left \lceil 2^{\mu} \right \rceil / L$}, xmin=0.2, xmax=1.1,
            xtick={ 0.3, 0.4, 0.5, 0.6, 0.7, 0.8 }, xticklabels = {0.3, 0.4, 0.5, 0.6, 0.7, 0.8},
            ylabel={\scriptsize probability density}, ymin=0,
            ytick={ 3,6,9,12 }, yticklabels={ 3,6,9,12 },
            ymajorgrids, major grid style={line width=.2pt, draw=black!20},
            legend style={at={ (0.95,0.65)}, anchor=east },
            legend cell align={left}
        ]
            \addplot[thick, fill=green!20, draw=green!80, smooth] table [x=bin, y=cnt, col sep=comma] {\loadedtableA} \closedcycle;
            \addlegendentry{\tiny $L\in [0, 200]$}

            \addplot[thick, fill=blue!50, draw=blue!60, smooth, opacity=0.6] table [x=bin, y=cnt, col sep=comma] {\loadedtableB} \closedcycle;
            \addlegendentry{\tiny $L\in [200, 350]$}

            \addplot[thick, fill=red!40, draw=red!60, smooth, opacity=0.6] table [x=bin, y=cnt, col sep=comma] {\loadedtableC} \closedcycle;
            \addlegendentry{\tiny $L\in [350,]$}

        \end{axis}
    \end{tikzpicture}   
    \caption{Distribution of low-rankness of BERT-base's embedding on IMDB dataset, measured by ratio $\left \lceil 2^{\mu} \right \rceil / L$. }
    \label{fig:lr:imdb}
\end{figure}
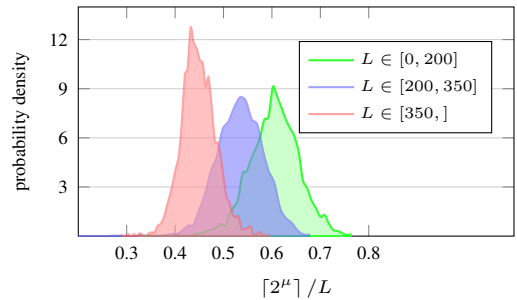

\section{Alternating Optimization}\label{sec:appx:alter}
\begin{algorithm}
    \SetAlgoLined
    \KwData{$r, X, \left\{ \ui{i}^0, \vi{i}^0 \right \}_{i=1}^r$}
    \KwResult{$\left\{ \ui{i}, \vi{i} \right \}_{i=1}^r$}
    \For{$i$ \textbf{in} $1,\cdots,r$}{
        \For{$j$ \textbf{in} $1,\cdots, 2$}{
            \tcc{Alternating optimization}
            $\ui{i}^{j} = \frac{X\cdot \vi{i}^{j-1}}{\left \| \vi{i}^{j-1}\right \|_F^2}$\;
            $\vi{i}^{j} = \frac{X^T\cdot \ui{i}^{j}}{\left \| \ui{i}^{j}\right \|_F^2}$\;
        }
    $\ui{i}, \vi{i}=\ui{i}^j, {\vi{i}^{j^T}}$\;
    $X = X - \ui{i}^j\cdot{\vi{i}^{j^T}}$\;
    }
\caption{Alternating Opt for SVD.}
\label{alg::lightSVD}
\end{algorithm}

\section{Finetune Hyperparameters}\label{sec:appx:hparam}
For BERT-base and Llama2 models, we conduct a grid search on learning rate (1e-5, 2e-5, 5e-5, 1e-4, 2e-4, 5e-4), and weight decay (1e-3, 5e-3, 1e-2, 5e-2).
Table \ref{tab:hparam:bert} and \ref{tab:hparam:llama} list the best hyperparameters found during fine-tuning.
\begin{table}[!htb]
    \centering
    \begin{tabular}{c|c|c|c|c}
    \toprule
        max len & batch size & epochs & $lr$ & $wd$ \\ \midrule
        512 & 32 & 20 & 5e-5 & 1e-2 \\
    \bottomrule
    \end{tabular}
    \caption{Finetuning hyperparameters for BERT-base on SST-2, Squad, and IMDB.}
    \label{tab:hparam:bert}
\end{table}

\begin{table}[!htb]
    \centering
    \begin{tabular}{c|c|c|c|c}
    \toprule
        max len & batch size & iters & $lr$ & $wd$ \\ \midrule
        2048 & 32 & 400 & 2e-4 & 1e-2 \\
    \bottomrule
    \end{tabular}
    \caption{Finetuning hyperparameters for Llama 2-7B/13B on MMLU and BoolQ.}
    \label{tab:hparam:llama}
\end{table}

\section{Prompt Format for MMLU and BoolQ}\label{sec:appx:prompt}

Table \ref{tab:mmlu:prompt} and \ref{tab:boolq:prompt} list the prompt format for MMLU and BoolQ dataset. 
\begin{table}[!htb]
    \centering
    \small
    \begin{tabular}{l}
    \toprule
        The following are multiple choice questions (with answers). \\
        One of the reasons that the government discourages \\ and regulates monopolies is that  \\
        A. producer surplus is lost and consumer surplus is gained. \\
        B. monopoly prices ensure productive efficiency but cost \\ society allocative efficiency. \\
        C. monopoly firms do not engage in significant research \\ and development. \\
        D. consumer surplus is lost with higher prices and lower \\ levels of output. \\
        Answer: \\
        C \\
    \bottomrule
    \end{tabular}
    \caption{MMLU prompt format}
    \label{tab:mmlu:prompt}
\end{table}

\begin{table}[!htb]
    \centering
    \small
    \begin{tabular}{l}
    \toprule
        Below is an instruction that describes a task. Write a response \\ that appropriately completes the request. \\
        \#\#\# Instruction: \\
        is harry potter and the escape from gringotts a roller coaster ride \\
        \#\#\# Input: \\
        Harry Potter and the Escape from Gringotts is an indoor steel \\ roller coaster ***\\ 
        \#\#\# Answer: \\
        Yes \\
    \bottomrule
    \end{tabular}
    \caption{BoolQ prompt format}
    \label{tab:boolq:prompt}
\end{table}


\end{document}